\documentclass[11pt,a4paper]{article}
\usepackage[hyperref]{eacl2021}
\usepackage{times}
\usepackage{latexsym}

\usepackage{microtype}

\usepackage{multirow}
\usepackage{latexsym}
\usepackage{booktabs}
\usepackage{courier}
\usepackage{amsmath}
\usepackage{subfig}
\usepackage{placeins}
\usepackage{tikz}
\usepackage{hyperref}

\aclfinalcopy 

\newcommand{\ra}[1]{\renewcommand{\arraystretch}{#1}}
\DeclareMathOperator*{\argmax}{argmax}

\makeatletter
\newcommand{\printfnsymbol}[1]{%
  \textsuperscript{\@fnsymbol{#1}}%
}
\makeatother

\title{Unification-based Reconstruction of Multi-hop Explanations for Science Questions}

\author{Marco Valentino\thanks{\ \ \ \texttt{equal contribution}}$~^{\dagger}$, Mokanarangan Thayaparan\printfnsymbol{1}$^{\dagger}$, Andr\'e Freitas$^{\dagger}$$^{\ddagger}$ \\  Department of Computer Science, University of Manchester, United Kingdom$^{\dagger}$ \\  Idiap Research Institute, Switzerland$^{\ddagger}$ \\ {\tt \{marco.valentino,mokanarangan.thayaparan,andre.freitas\}} \\ {\tt @manchester.ac.uk} \\}

\begin{document}
\maketitle
\begin{abstract}
This paper presents a novel framework for reconstructing \emph{multi-hop explanations} in science Question Answering (QA). While existing approaches for multi-hop reasoning build explanations considering each question in isolation, we propose a method to leverage \emph{explanatory patterns} emerging in a corpus of scientific explanations. Specifically, the framework ranks a set of atomic facts by integrating lexical relevance with the notion of \emph{unification power}, estimated analysing explanations for similar questions in the corpus. 

An extensive evaluation is performed on the Worldtree corpus, integrating k-NN clustering and Information Retrieval (IR) techniques. We present the following conclusions: (1) The proposed method achieves results competitive with Transformers, yet being orders of magnitude faster, a feature that makes it scalable to large explanatory corpora (2) The unification-based mechanism has a key role in reducing semantic drift, contributing to the reconstruction of many hops explanations (6 or more facts) and the ranking of complex inference facts (+12.0 Mean Average Precision) (3) Crucially, the constructed explanations can support downstream QA models, improving the accuracy of BERT by up to 10\% overall.
\end{abstract}

\section{Introduction}
Answering \emph{multiple-choice science questions} has become an established benchmark for testing natural language understanding and complex reasoning in Question Answering (QA)~\cite{khot2019qasc,clark2018think,mihaylov2018can}.
In parallel with other NLP research areas, a crucial requirement emerging in recent years is \emph{explainability}~\cite{thayaparan2020survey,miller2019explanation,biran2017explanation,ribeiro2016should}. To boost automatic methods of inference, it is necessary not only to measure the performance on answer prediction, but also the ability of a QA system to provide explanations for the underlying reasoning process. 

The need for explainability and a quantitative methodology for its evaluation have conducted to the creation of shared tasks on \emph{explanation reconstruction}~\cite{jansen2019textgraphs} using corpora of explanations such as Worldtree~\cite{jansen2018worldtree,jansen2016s}.
Given a science question, explanation reconstruction consists in regenerating the gold explanation that supports the correct answer through the combination of a series of atomic facts. While most of the existing benchmarks for multi-hop QA require the composition of only 2 supporting sentences or paragraphs (e.g. QASC \cite{khot2019qasc}, HotpotQA \cite{yang2018hotpotqa}), the explanation reconstruction task requires the aggregation of an average of 6 facts (and as many as $\approx$20), making it particularly hard for  multi-hop reasoning models. Moreover, the structure of the explanations affects the complexity of the reconstruction task. Explanations for science questions are typically composed of two main parts: a grounding part, containing knowledge about concrete concepts in the question, and a core scientific part, including general scientific statements and laws.

Consider the following question and answer pair from Worldtree \cite{jansen2018worldtree}:
\begin{itemize}
\item $q$: what is an example of a \textbf{force} producing heat? \\ $a$: two \textbf{sticks} getting warm when \textbf{rubbed together}.
\end{itemize}
An explanation that justifies $a$ is composed using the following sentences from the corpus: \emph{($f_1$) a \textbf{stick} is a kind of object}; \emph{($f_2$) to \textbf{rub together} means to move against}; \emph{($f_3$) friction is a kind of \textbf{force}}; \emph{($f_4$) friction occurs when two objects' surfaces move against each other}; \emph{($f_5$) friction causes the temperature of an object to increase}.
The explanation contains a set of concrete sentences that are conceptually connected with $q$ and $a$ ($f_1$,$f_2$ and $f_3$), along with a set of abstract facts that require multi-hop inference ($f_4$ and $f_5$).

Previous work has shown that constructing long explanations is challenging due to \emph{semantic drift} -- i.e. the tendency of composing out-of-context inference chains as the number of hops increases~\cite{khashabi2019capabilities,fried2015higher}.
While existing approaches build explanations considering each question in isolation \cite{khashabi2018question,khot2017answering}, we hypothesise that semantic drift can be tackled by leveraging \emph{explanatory patterns} emerging in clusters of similar questions. 

In Science, a given statement is considered explanatory to the extent it performs \emph{unification}~\cite{friedman1974explanation,kitcher1981explanatory,kitcher1989explanatory}, that is showing how a set of initially disconnected phenomena are the expression of the same regularity. An example of unification is Newton's law of universal gravitation, which \emph{unifies} the motion of planets and falling bodies on Earth showing that \emph{all} bodies with mass obey the same law. Since the explanatory power of a given statement depends on the number of unified phenomena, highly explanatory facts tend to create \emph{unification patterns} -- i.e. similar phenomena require similar explanations. Coming back to our example, we hypothesise that the relevance of abstract statements requiring multi-hop inference, such as $f_4$ (\emph{``friction occurs when two objects' surfaces move against each other''}), can be estimated by taking into account the unification power.

Following these observations, we present a framework that ranks atomic facts through the combination of two scoring functions: 

\begin{itemize}
    \item A \emph{Relevance Score (RS)}  that represents the lexical relevance of a given fact.
    \item A \emph{Unification Score (US)} that models the explanatory power of a fact according to its frequency in explanations for similar questions. 
\end{itemize}

An extensive evaluation is performed on the Worldtree corpus \cite{jansen2018worldtree,jansen2019textgraphs}, adopting a combination of k-NN clustering and Information Retrieval (IR) techniques. We present the following conclusions:

\begin{enumerate}
    \item Despite its simplicity, the proposed method achieves results competitive with Transformers \cite{das2019chains,chia2019red}, yet being orders of magnitude faster, a feature that makes it scalable to large explanatory corpora.
    \item We empirically demonstrate the key role of the unification-based mechanism in the reconstruction of many hops explanations (6 or more facts) and explanations requiring complex inference (+12.0 Mean Average Precision). 
    \item Crucially, the constructed explanations can support downstream question answering models, improving the accuracy of BERT \cite{devlin2019bert} by up to 10\% overall. 
\end{enumerate}

To the best of our knowledge, we are the first to propose a method that leverages unification patterns for the reconstruction of multi-hop explanations, and empirically demonstrate their impact on semantic drift and downstream question answering.

\section{Related Work}

\paragraph{Explanations for Science Questions.}
Reconstructing explanations for science questions can be reduced to a multi-hop inference problem, where multiple pieces of evidence have to be aggregated to arrive at the final answer~\cite{thayaparan2020survey,khashabi2018question,khot2017answering,jansen2017framing}.
Aggregation methods based on lexical overlaps and explicit constraints suffer from \emph{semantic drift} ~\cite{khashabi2019capabilities,fried2015higher} -- i.e. the tendency of composing spurious inference chains leading to wrong conclusions. 

One way to contain semantic drift is to leverage common explanatory patterns in explanation-centred corpora~\cite{jansen2018worldtree}. 
Transformers~\cite{das2019chains,chia2019red} represent the state-of-the-art for explanation reconstruction in this setting \cite{jansen2019textgraphs}. However, these models require high computational resources that prevent their applicability to large corpora. On the other hand, approaches based on IR techniques are readily scalable. The approach described in this paper preserves the scalability of IR methods, obtaining, at the same time, performances competitive with Transformers. Thanks to this feature, the framework can be flexibly applied in combination with downstream question answering models.

Our findings are in line with previous work in different QA settings~\cite{rajani2019explain,yadav2019quick}, which highlights the positive impact of explanations and supporting facts on the final answer prediction task. 

In parallel with Science QA, the development of models for explanation generation is being explored in different NLP tasks,
ranging from open domain question answering \cite{yang2018hotpotqa,thayaparan2019identifying}, to
textual entailment \cite{camburu2018snli} and natural language premise selection \cite{ferreira2020premise,ferreira2020natural}.

\paragraph{Scientific Explanation and AI.}
The field of Artificial Intelligence has been historically inspired by models of explanation in Philosophy of Science~\cite{thagard2008models}. The deductive-nomological model proposed by Hempel~\cite{hempel1965aspects} constitutes the philosophical foundation for explainable models based on logical deduction, such as Expert Systems~\cite{lacave2004review,wick1992reconstructive} and Explanation-based Learning~\cite{mitchell1986explanation}. Similarly, the inherent relation between explanation and causality~\cite{woodward2005making,salmon1984scientific} has inspired computational models of causal inference~\cite{pearl2009causality}. 
The view of explanation as unification~\cite{friedman1974explanation,kitcher1981explanatory,kitcher1989explanatory} is closely related to Case-based reasoning~\cite{kolodner2014case,sormo2005explanation,de2005retrieval}. 
In this context, analogical reasoning plays a key role in the process of reusing abstract patterns for explaining new phenomena~\cite{thagard1992analogy}. Similarly to our approach, Case-based reasoning applies this insight to construct solutions for novel problems by retrieving, reusing and adapting explanations for known cases solved in the past. 

\begin{figure*}
\centering
\includegraphics[width=\textwidth]{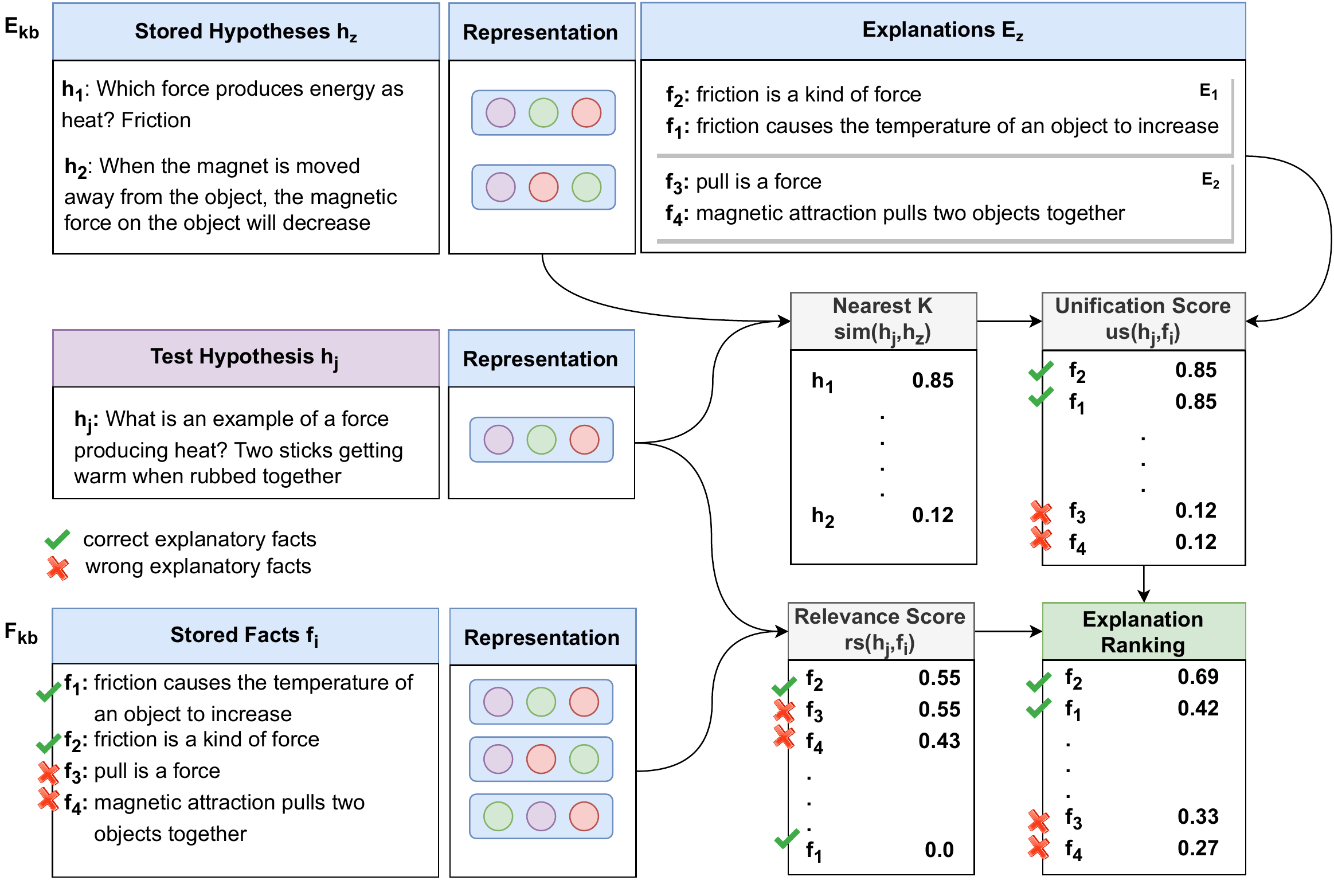}
\caption{Overview of the Unification-based framework for explanation reconstruction.}
\label{fig:approach}
\end{figure*}

\section{Explanation Reconstruction as a Ranking Problem}

A multiple-choice science question $Q = \{q, C\}$  is a tuple composed by a question $q$  and a set of candidate answers $C =\{c_1,c_2,\ldots,c_n\}$. Given an hypothesis $h_j$ defined as the concatenation of $q$  with a candidate answer $c_j \in C$, the task of explanation reconstruction consists in selecting a set of atomic facts from a knowledge base $E_j = \{ f_1,f_2,\ldots,f_n\}$  that support and justify $h_j$.

In this paper, we adopt a methodology that relies on the existence of a corpus of explanations. A corpus of explanations is composed of two distinct knowledge sources:

\begin{itemize}
    \item A primary knowledge base,  \emph{Facts KB} ($F_{kb}$), defined as a collection of sentences $F_{kb} = \{f_1,f_2,\ldots,f_n\}$ encoding the general world knowledge necessary to answer and explain science questions. A fundamental and desirable characteristic of $F_{kb}$ is \emph{reusability} -- i.e. each of its facts $f_i$ can be potentially reused to compose explanations for multiple questions
    \item A secondary knowledge base,  \emph{Explanation KB} ($E_{kb}$), consisting of a set of tuples $E_{kb} = \{(h_1,E_1),(h_2, E_2), \ldots,(h_m, E_m)\}$, each of them connecting a true hypothesis $h_j$ to its corresponding explanation $E_{j} = \{f_1,f_2, \ldots, f_{k}\} \subseteq F_{kb}$. An explanation $ E_j \in E_{kb}$ is therefore a composition of facts belonging to $F_{kb}$.
\end{itemize}

In this setting, the explanation reconstruction task for an unseen hypothesis $h_j$ can be modelled as a ranking problem~\cite{jansen2019textgraphs}. Specifically, given an hypothesis $h_j$ the algorithm to solve the task is divided into three macro steps:
\begin{enumerate}
    \item Computing an explanatory score $s_i = e(h_j,f_i)$ for each fact $f_i \in F_{kb}$ with respect to $h_j$
    \item Producing an ordered set $Rank(h_j) = \{f_1,\ldots,f_k,f_{k+1},\ldots,f_n \text{ }|\text{ } s_k \geq s_{k+1}\} \subseteq F_{kb}$
    \item Selecting the top $k$ elements belonging to $Rank(h_j)$ and interpreting them as an explanation for $h_j$; $E_j = topK(Rank(h_j))$.
\end{enumerate}

\subsection{Modelling Explanatory Relevance}

We present an approach for modelling $e(h_j,f_i)$ that is guided by the following research hypotheses: 

\begin{itemize}
    \item \textbf{RH1}: Scientific explanations are composed of a set of concrete facts connected to the question, and a set of abstract statements expressing general scientific laws and regularities.
    \item \textbf{RH2}: Concrete facts tend to share key concepts with the question and can therefore be effectively ranked by IR techniques based on lexical relevance.
    \item \textbf{RH3}: General scientific statements tend to be abstract and therefore difficult to rank by means of shared concepts. However, due to explanatory unification, core scientific facts tend to be frequently reused across similar questions. We hypothesise that the explanatory power of a fact $f_i$ for a given hypothesis $h_j$ is proportional to the number of times $f_i$ explains similar hypotheses.
\end{itemize}

To formalise these research hypotheses, we model the explanatory scoring function $e(h_j, f_i)$ as a combination of two components:
\begin{equation}
\small
    e(h_j, f_i) = \lambda_1 rs(h_j, f_i) + (1-\lambda_1) us(h_j, f_i)
    \label{eq:model_combination}
\end{equation}
Here, $rs(h_j, f_i)$ represents a lexical Relevance Score (RS) assigned to $f_i \in F_{kb}$ with respect to $h_j$, while $us(h_j, f_i)$ represents the Unification Score (US) of $f_i$ computed over $E_{kb}$  as follows:
\begin{equation}
\small
us(h_j,f_i) = \sum_{(h_z,E_z) \in kNN(h_j)} sim(h_j,h_z) in(f_i,E_z)
\label{eq:unification_score}
\end{equation}  
\begin{equation}
\small
    in(f_i,E_z) =
        \begin{cases}
            1 & \text{if } f_i \in E_{z}\\
            0 & \text{otherwise}
        \end{cases}
\end{equation}

$kNN(h_j) = \{ (h_1, E_1), \ldots (h_k, E_k)\} \subseteq E_{kb}$ is the set of k-nearest neighbours of $h_j$ belonging to $E_{kb}$ retrieved according to a similarity function $sim(h_j,h_z)$. On the other hand, $in(f_i,E_z)$ verifies whether the fact $f_i$ belongs to the explanation $E_z$ for the hypothesis $h_z$.

In the formulation of Equation~\ref{eq:unification_score} we aim to capture two main aspects related to our research hypotheses: 
\begin{enumerate}
    \item The more a fact $f_i$ is reused for explanations in $E_{kb}$, the higher its explanatory power and therefore its Unification Score; 
    \item The Unification Score of a fact $f_i$ is proportional to the similarity between the hypotheses in $E_{kb}$ that are explained by $f_i$  and the unseen hypothesis ($h_j$) we want to explain. 
\end{enumerate}
Figure \ref{fig:approach} shows a schematic representation of the Unification-based framework.

\begin{table*}
    \small
    \centering
    \ra{1}
    \begin{tabular}{@{}p{5cm}p{5cm}ccc@{}}
        \toprule
         \multirow{2}{*}{\textbf{Model}} & \multirow{2}{*}{\textbf{Approach}} &
         \multirow{2}{*}{\textbf{Trained}} &
         \multicolumn{2}{c}{\textbf{MAP}}\\
         \cmidrule{4-5}
         & & & Test & Dev  \\ 
         \midrule
       \textbf{Transformers} & & & \\
       \midrule
       Das et al.~\shortcite{das2019chains} & BERT re-ranking with inference chains & Yes & \textbf{56.3} & \textbf{58.5} \\
        Chia et al.~\shortcite{chia2019red} & BERT re-ranking with gold IR scores & Yes & 47.7 &  50.9 \\
        Banerjee~\shortcite{banerjee2019asu} & BERT iterative re-ranking & Yes & 41.3 &  42.3 \\
         \midrule
       \textbf{IR with re-ranking} & & & \\
       \midrule
         Chia et al.~\shortcite{chia2019red} & Iterative BM25 & \textbf{No} & 45.8 &  49.7 \\
         \midrule
       \textbf{One-step IR} & & & \\
       \midrule
        BM25 & BM25 Relevance Score & \textbf{No} & 43.0 &  46.1 \\
        TF-IDF & TF-IDF Relevance Score & \textbf{No} & 39.4 &  42.8\\
        \midrule
       \textbf{Feature-based} & & & \\
       \midrule
        D’Souza et al.\shortcite{d2019team} & Feature-rich SVM ranking + Rules  & Yes & 39.4 &  44.4 \\
        D’Souza et al.~\shortcite{d2019team} & Feature-rich SVM ranking & Yes & 34.1 &  37.1\\ \bottomrule
        \toprule
       \textbf{Unification-based Reconstruction} & & & \\
       \midrule
         RS + US (Best) & Joint Relevance and Unification Score & \textbf{No} & \textbf{50.8} &  \textbf{54.5} \\
         US (Best) & Unification Score & \textbf{No} & 22.9 &  21.9 \\
        \bottomrule
    \end{tabular}
    \caption{Results on test and dev set and comparison with state-of-the-art approaches. The column \textbf{trained} indicates whether the model requires an explicit training session on the explanation reconstruction task.}
    \label{tab:compare_approaches_overall}
\end{table*}

\section{Empirical Evaluation}

We carried out an empirical evaluation on the Worldtree corpus ~\cite{jansen2018worldtree}, a subset of the ARC dataset~\cite{clark2018think} that includes explanations for science questions. The corpus provides an explanatory knowledge base ($F_{kb}$ and $E_{kb}$) where each explanation in $E_{kb}$ is represented as a set of lexically connected sentences describing how to arrive at the correct answer. The science questions in the Worldtree corpus are split into \emph{training-set}, \emph{dev-set}, and \emph{test-set}. The gold explanations in the \emph{training-set} are used to form the Explanation KB ($E_{kb}$), while the gold explanations in \emph{dev} and \emph{test} set are used for \emph{evaluation purpose only}. The corpus groups the explanation sentences belonging to $E_{kb}$ into three explanatory roles: \textit{grounding}, \textit{central} and \textit{lexical glue}.

Consider the example in Figure~\ref{fig:approach}. To support $q$ and $c_j$ 
the system has to retrieve the scientific facts describing how friction occurs and produces heat across objects. 
The corpus classifies these facts ($f_3, f_4)$ as \textit{central}. \textit{Grounding} explanations like \emph{``stick is a kind of object''} ($f_1$) link question and answer to the central explanations. \textit{Lexical glues} such as \emph{``to rub; to rub together means to mover against''} ($f_2$) are used to fill lexical gaps between sentences.
Additionally, the corpus divides the facts belonging to $F_{kb}$ into three inference categories: \textit{retrieval type}, \textit{inference supporting type}, and \textit{complex inference type}. Taxonomic knowledge and properties such as \emph{``stick is a kind of object"} ($f_1$)  and \emph{``friction is a kind of force"} ($f_5$) are classified as \textit{retrieval type}. Facts describing actions, affordances, and requirements such as \emph{``friction occurs when two object's surfaces move against each other"} ($f_3$) are grouped under the \textit{inference supporting types}. Knowledge about causality, description of processes and if-then conditions such as \emph{``friction causes the temperature of an object to increase''} ($f_4$) is classified as \textit{complex inference}.

We implement Relevance and Unification Score adopting TF-IDF/BM25 vectors and cosine similarity function (i.e. $1-cos(\vec{x},\vec{y})$). 
Specifically, The RS model uses TF-IDF/BM25 to compute the relevance function for each fact in $F_{kb}$ (i.e. $rs(h_j, f_i)$ function in Equation~\ref{eq:model_combination}) while the US model adopts TF-IDF/BM25 to assign similarity scores to the hypotheses in $E_{kb}$ (i.e. $sim(h_j,h_z)$ function in Equation~\ref{eq:unification_score}). For reproducibility, the code is available at the following url: \url{https://github.com/ai-systems/unification_reconstruction_explanations}. Additional details can be found in the supplementary material.

\subsection{Explanation Reconstruction}

In line with the shared task~\cite{jansen2019textgraphs}, the  performances of the models are evaluated via Mean Average Precision (MAP) of the explanation ranking produced for a given question $q_j$ and its correct answer $a_j$. 

Table~\ref{tab:compare_approaches_overall} illustrates the score achieved by our best implementation compared to state-of-the-art approaches in the literature.
Previous approaches are grouped into four categories: \textit{Transformers}, \textit{Information Retrieval with re-ranking}, \textit{One-step Information Retrieval}, and \textit{Feature-based models}.

\begin{table*}[t]
     \centering
    \small
    \ra{1}
    \subfloat[Explanatory roles.    \label{tab:compare_approaches_explanations_type}]{\resizebox{0.5\textwidth}{!}{
    \begin{tabular}{@{}p{4cm}cccc@{}}
    \toprule
         \multirow{2}{*}{\textbf{Model}} & \multicolumn{4}{c}{\textbf{MAP}}  \\
         \cmidrule{2-5}
         & All & Central & Grounding & Lexical Glue\\
         \midrule
         RS TF-IDF & 42.8 & 43.4 &  25.4 & 8.2 \\
         RS BM25 & 46.1 &  46.6 &  23.3 & 10.7 \\
         \midrule
         US TF-IDF & 21.6 & 16.9 &  22.0 & 13.4 \\
         US BM25 & 21.9 & 18.1 &  16.7 & 15.0 \\
         \midrule
         RS TF-IDF + US TF-IDF & 48.5 & 46.4 & \textbf{32.7} & 11.7\\
         RS TF-IDF + US BM25 & 50.7 & 48.6 & 30.42 & 13.4\\
         RS BM25 + US TF-IDF & 51.9 & 48.2 & 31.7 & 14.8\\
         RS BM25 + US BM25 & \textbf{54.5} & \textbf{51.7} & 27.3 & \textbf{16.7}\\
         \bottomrule
    \end{tabular}
    }}
    \hfill
    \subfloat[Lexical overlaps with the hypothesis. \label{tab:compare_approaches_overlaps}]{\resizebox{0.45\textwidth}{!}{
    \begin{tabular}{@{}p{4cm}ccc@{}}
    \toprule
         \multirow{2}{*}{\textbf{Model}} & \multicolumn{3}{c}{\textbf{MAP}} \\
         \cmidrule{2-4}
         & 1+ Overlaps & 1 Overlap & 0 Overlaps\\
         \midrule
         RS TF-IDF & 57.2 & 33.6 &  7.1 \\
         RS BM25 & 62.2 &  37.1 &  7.1\\
         \midrule
         US TF-IDF & 17.37 & 18.0 & 12.5 \\
         US BM25 & 18.1 & 18.1 &  \textbf{13.1}\\
         \midrule
         RS TF-IDF + US TF-IDF & 60.2 & 38.4 & 9.0\\
         RS TF-IDF + US BM25 & 62.5 & 39.5 & 9.6\\
         RS BM25 + US TF-IDF & 61.3 & 40.6 & 11.0\\
         RS BM25 + US BM25 & \textbf{64.8} & \textbf{41.9} & 11.2\\
         \bottomrule
    \end{tabular}}}
    \hfill
    \subfloat[Inference types.    \label{tab:compare_approaches_inference_type}]{\resizebox{0.55\textwidth}{!}{
    \begin{tabular}{@{}p{4cm}ccc@{}}
    \toprule
         \multirow{2}{*}{\textbf{Model}} & \multicolumn{3}{c}{\textbf{MAP}}  \\
         \cmidrule{2-4}
          & Retrieval & Inference-supporting & Complex Inference\\
         \midrule
         RS TF-IDF & 33.5 &  34.7 & 21.8 \\
         RS BM25 &  36.0 &  36.1 & 24.8 \\
         \midrule
         US TF-IDF & 17.6 &  12.8 & 19.5 \\
         US BM25 & 16.8 &  13.2 & 20.9 \\
         \midrule
         RS TF-IDF + US TF-IDF & 38.3 & 33.2 & 30.2\\
         RS TF-IDF + US BM25 & 40.0 & 35.6 & 33.3\\
         RS BM25 + US TF-IDF & 40.5 & 33.6 & 33.4\\
         RS BM25 + US BM25 & \textbf{40.6} & \textbf{38.3} & \textbf{36.8}\\
         \bottomrule
    \end{tabular}}}
    \caption{Detailed analysis of the performance (dev-set) by breaking down the gold explanatory facts according to their explanatory role (2.a), number of lexical overlaps with the question (2.b) and inference type (2.c).}
\end{table*}

\paragraph{Transformers.} This class of approaches employs the gold explanations in the corpus to train a BERT language model~\cite{devlin2019bert}. 
The best-performing system~\cite{das2019chains} adopts a multi-step retrieval strategy. In the first step, it returns the top K sentences ranked by a TF-IDF model. In the second step, BERT is used to re-rank the paths composed of all the facts that are within 1-hop from the first retrieved set. Similarly, other approaches adopt BERT to re-rank each fact individually~\cite{banerjee2019asu,chia2019red}. 

Although the best model achieves state-of-the-art results in explanation reconstruction, these approaches are computationally expensive, being limited by the application of a pre-filtering step to contain the space of candidate facts. Consequently, these systems do not scale with the size of the corpus. We estimated that the best performing model \cite{das2019chains} takes $\approx 10$ hours to run on the whole test set (1240 questions) using 1 Tesla 16GB V100 GPU. 

Comparatively, our model constructs explanations for all the questions in the test set in $\approx$ 30 seconds, without requiring the use of GPUs ($<1$ second per question). This feature makes the Unification-based Reconstruction suitable for large corpora and downstream question answering models (as shown in Section \ref{sec:answ_sel}). Moreover, our approach does not require any explicit training session on the explanation regeneration task, with significantly reduced number of parameters to tune. Along with scalability, the proposed approach achieves nearly state-of-the-art results (50.8/54.5 MAP). Although we observe lower performance when compared to the best-performing approach (-5.5/-4.0 MAP), the joint RS + US model outperforms two BERT-based models~\cite{chia2019red,banerjee2019asu} on both test and dev set by 3.1/3.6 and 9.5/12.2 MAP respectively.

\paragraph{Information Retrieval with re-ranking.} Chia et al.~\shortcite{chia2019red} describe a multi-step, iterative re-ranking model based on BM25. The first step consists in retrieving the explanation sentence that is most similar to the question adopting BM25 vectors. During the second step, the BM25 vector of the question is updated by aggregating it with the retrieved explanation sentence vector through a \texttt{max} operation. The first and second steps are repeated for $K$ times. Although this approach uses scalable IR techniques, it relies on a multi-step retrieval strategy. Besides, the RS + US model outperforms this approach on both test and dev set by 5.0/4.8 MAP respectively.

\begin{figure*}
\centering
\subfloat[MAP vs Explanation length.\label{fig:map_exp_length}]{\includegraphics[width=0.48\textwidth]{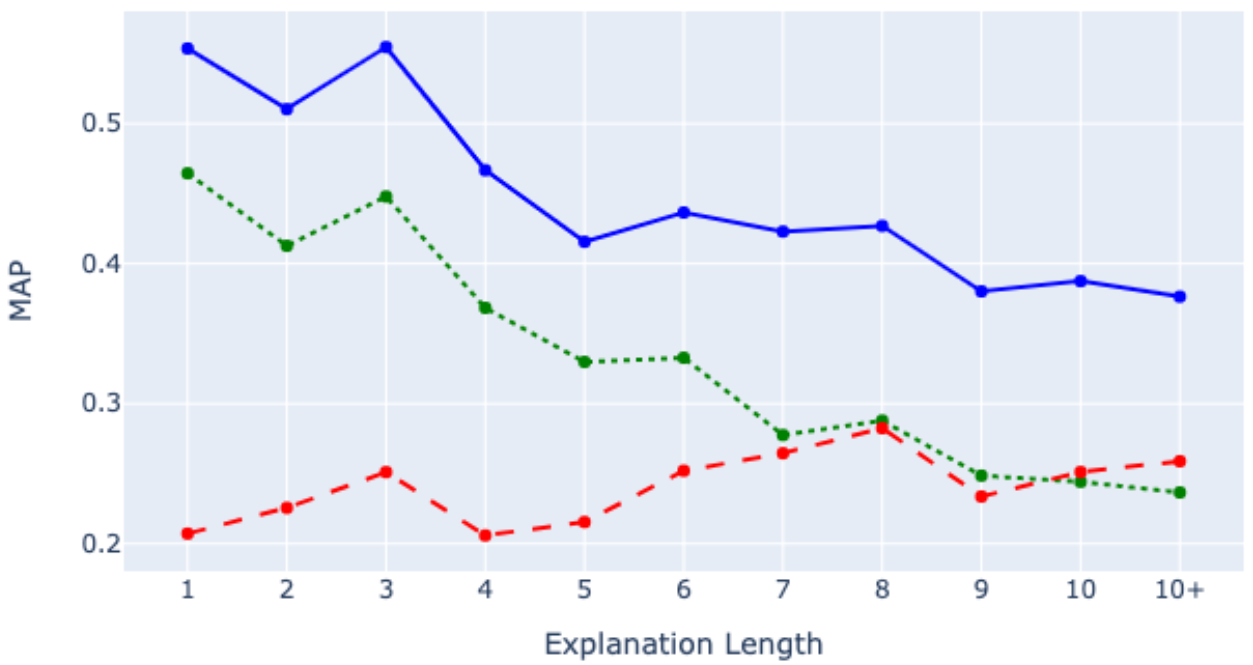}}
\hfill
\subfloat[Precision@K.\label{fig:precision_k}]{\includegraphics[width=0.48\textwidth]{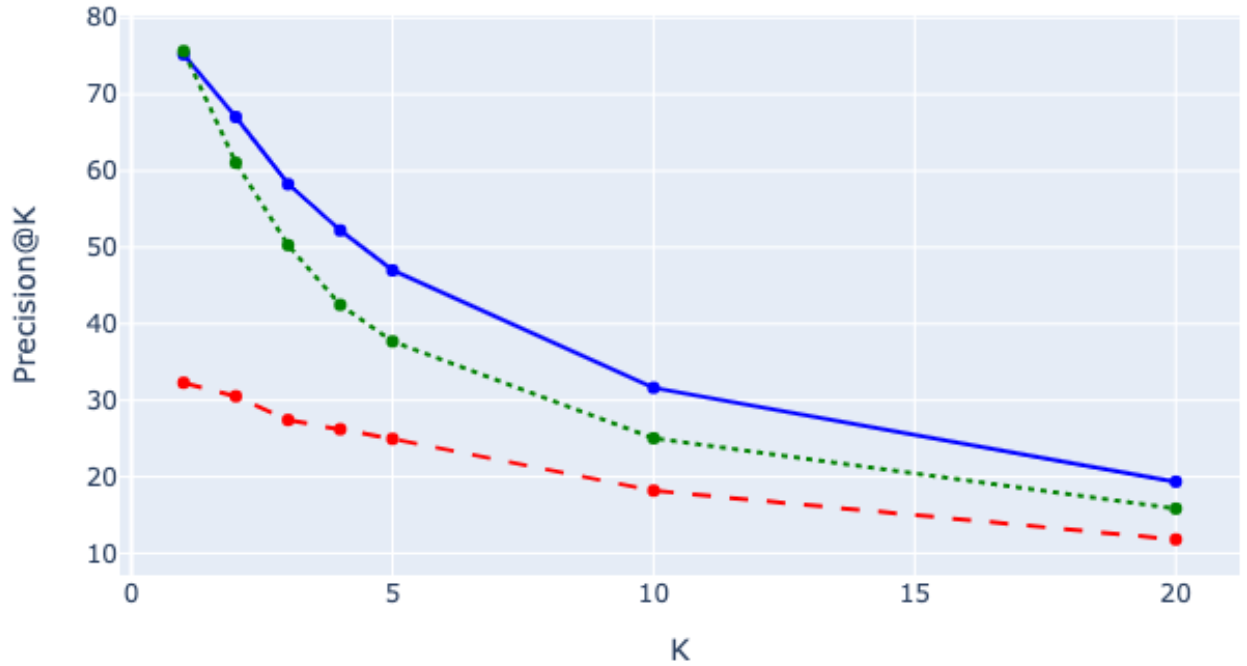}}
\caption{Impact of the Unification Score on semantic drift (3.a) and precision (3.b). RS + US ({\color{blue}Blue Straight}), RS ({\color{green}Green Dotted}), US ({\color{red}Red Dashed}).}
\end{figure*}

\begin{figure}
        \centering
        \includegraphics[width=\columnwidth]{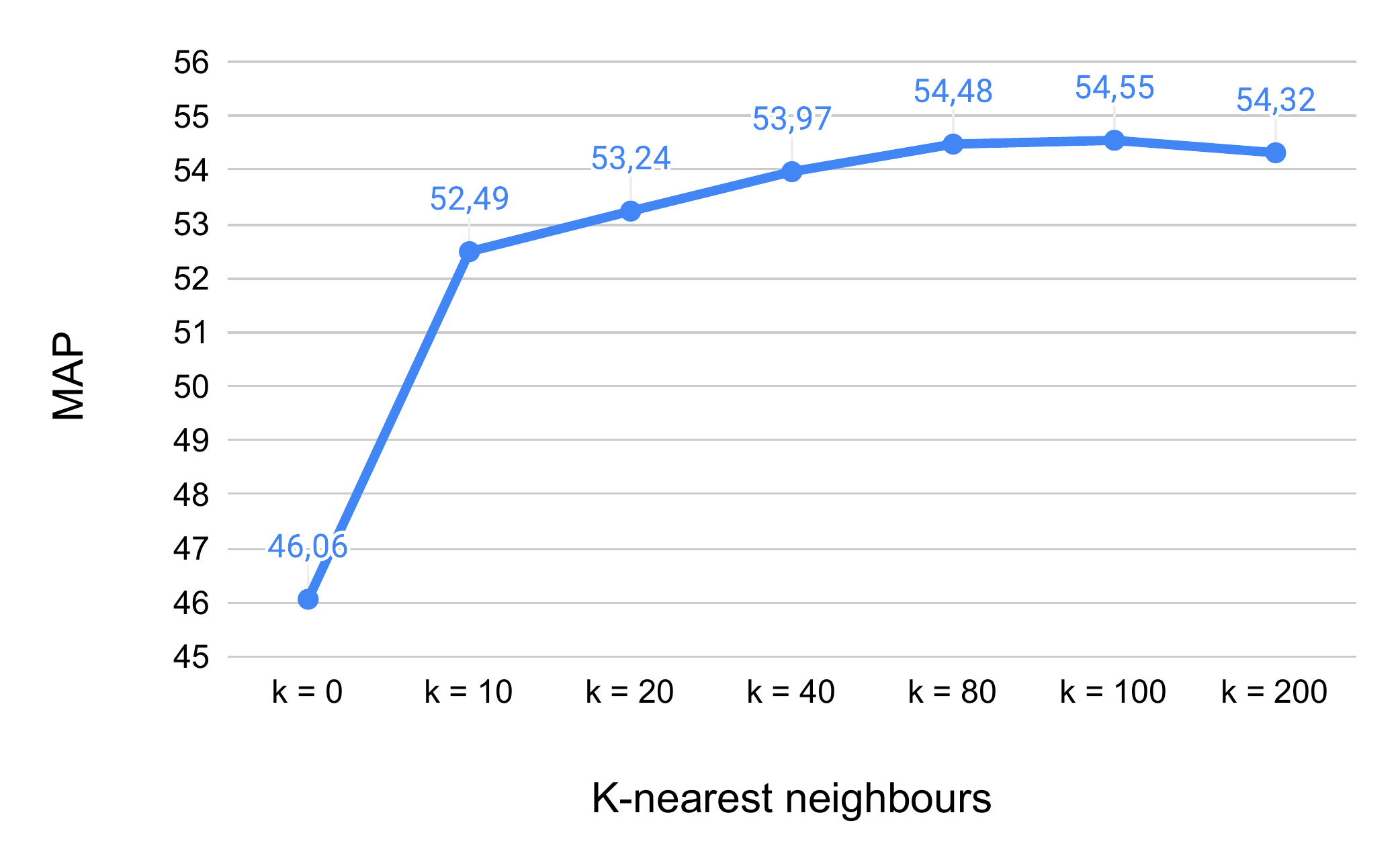}
        \caption{Impact of the k-NN clustering on the final MAP score. The value $k$ represents the number of similar hypotheses considered for the Unification Score.}
        \label{fig:impact_knn}
\end{figure}

\paragraph{One-step Information Retrieval.} We compare the RS + US model with two IR baselines. The baselines adopt TF-IDF and BM25 to compute the Relevance Score only -- i.e. the $us(q, c_j, f_i)$ term in Equation 1 is set to 0 for each fact $f_i \in F_{kb}$. In line with previous IR literature~\cite{robertson2009probabilistic}, BM25 leads to better performance than TF-IDF. While these approaches share similar characteristics, the combined RS + US model outperforms both RS BM25 and RS TF-IDF on test and dev-set by 7.8/8.4 and 11.4/11.7 MAP. Moreover, the joint RS + US model improves the performance of the US model alone by 27.9/32.6 MAP. These results outline the complementary aspects of Relevance and Unification Score. We provide a detailed analysis by performing an ablation study on the dev-set (Section \ref{sec:ablation}).   

\paragraph{Feature-based models.} D’Souza  et  al.~\shortcite{d2019team} propose an approach based on a learning-to-rank paradigm. The model extracts a set of features based on overlaps and coherence metrics between questions and explanation sentences. These features are then given in input to a SVM ranker module. While this approach scales to the whole corpus without requiring any pre-filtering step, it is significantly outperformed by the RS + US model on both test and dev set by 16.7/17.4 MAP respectively.

\subsection{Explanation Analysis}
\label{sec:ablation}
We present an ablation study with the aim of understanding the contribution of each sub-component to the general performance of the joint RS + US model (see Table \ref{tab:compare_approaches_overall}). 
To this end, a detailed evaluation on the development set of the Worldtree corpus is carried out, analysing the performance in reconstructing explanations of different types and complexity. We compare the joint model (RS + US) with each individual sub-component (RS and US alone).
In addition, a set of qualitative examples are analysed to provide additional insights on the complementary aspects captured by Relevance and Unification Score.

\begin{table*}[t]
    \centering
    \small
    \ra{1}
    \resizebox{\textwidth}{!}{\begin{tabular}{@{}p{3.5cm}p{2.5cm}p{4cm}p{5.5cm}cc@{}}
    \toprule
         \multirow{2}{*}{\textbf{Question}} &
         \multirow{2}{*}{\textbf{Answer}} &
         \multirow{2}{*}{\textbf{Explanation Fact}} &
         \multirow{2}{*}{\textbf{Most Similar Hypotheses in $E_{kb}$}} &
         \multirow{2}{*}{\textbf{RS}} & \multirow{2}{*}{\textbf{RS + US}} \\\\
         \midrule
         If you bounce a rubber ball on the floor  it goes up and then comes down. What \textbf{causes} the ball to come down? &  \textbf{gravity} & \textbf{gravity}; gravitational force \textbf{causes} objects that have mass; substances to be pulled down; to fall on a planet & (1) A ball is tossed up in the air and it comes back down. The ball comes back down because of - gravity (2) A student drops a ball. Which force causes the ball to fall to the ground? - gravity & \#36 & \#2 ($\uparrow$34)\\
         \midrule
         Which \textbf{animals} would most likely be helped by flood in a coastal area? & alligators & as water increases in an environment, the population of aquatic \textbf{animals} will increase & (1) Where would animals and plants be most affected by a flood? - low areas (2) Which change would most likely increase the number of salamanders? - flood & \#198 & \#57 ($\uparrow$141)\\
         \midrule
         What is an example of a force producing heat? & two sticks getting warm when rubbed together & friction causes the temperature of an object to increase & (1) Rubbing sandpaper on a piece of wood produces what two types of energy? - sound and heat (2) Which force produces energy as heat? - friction & \#1472 & \#21 ($\uparrow$1451)\\
         \bottomrule
    \end{tabular}}
    \caption{Impact of the Unification Score on the ranking of scientific facts with increasing complexity.}
    \label{tab:qualitative_examples}
\end{table*}

\paragraph{Explanatory categories.}
Given a question $q_j$ and its correct answer $a_j$, we classify a fact $f_i$ belonging to the gold explanation $E_j$ according to its explanatory role  (\emph{central, grounding, lexical glue}) and inference type (\emph{retrieval, inference-supporting and complex inference}).
In addition, three new categories are derived from the number of overlaps between $f_i$ and the concatenation of $q_j$ with $a_j$ ($h_j$) computed by considering nouns, verbs, adjectives and adverbs (1+ overlaps, 1 overlap, 0 overlaps).

Table 2 reports the MAP score for each of the described categories.
Overall, the best results are obtained by the BM25 implementation of the joint model (RS BM25 + US BM25) with a MAP score of 54.5. Specifically, RS BM25 + US BM25 achieves a significant improvement over both RS BM25 (+8.5 MAP) and US BM25 (+32.6 MAP) baselines. Regarding the explanatory roles (Table \ref{tab:compare_approaches_explanations_type}), the joint TF-IDF implementation shows the best performance in the reconstruction of \emph{grounding} explanations (32.7 MAP). On the other hand, a significant improvement over the RS baseline is obtained by RS BM25 + US BM25 on both \emph{lexical} glues and \emph{central} explanation sentences (+6.0 and +5.6 MAP over RS BM25).

Regarding the lexical overlaps categories (Table \ref{tab:compare_approaches_overlaps}), we observe a steady improvement for all the combined RS + US models over the respective RS baselines. Notably, the US models achieve the best performance on the 0 overlaps category, which includes the most challenging facts for the RS models. The improved ability to rank abstract explanatory facts contributes to better performance for the joint models (RS + US) in the reconstruction of explanations that share few terms with question and answer (\emph{1 Overlap} and \emph{0 Overlaps} categories). This characteristic leads to an improvement of 4.8 and 4.1 MAP for the RS BM25 + US BM25 model over the RS BM25 baseline. 

Similar results are achieved on the inference types categories (Table \ref{tab:compare_approaches_inference_type}).
Crucially, the largest improvement is observed for \emph{complex inference} sentences where RS BM25 + US BM25 outperforms RS BM25 by 12.0 MAP, confirming the decisive contribution of the Unification Score to the ranking of complex scientific facts.

\paragraph{Semantic drift.} 
Science questions in the Worldtree corpus require an average of six facts in their explanations ~\cite{jansen2016s}. Long explanations typically include sentences that share few terms with question and answer, increasing the probability of semantic drift. Therefore, to test the impact of the Unification Score on the robustness of the model, we measure the performance in the reconstruction of many-hops explanations. 

Figure~\ref{fig:map_exp_length} shows the change in MAP score for the RS + US, RS and US models (BM25) with increasing explanation length. The fast drop in performance for the Relevance Score reflects the complexity of the task. This drop occurs because the RS model is not able to rank abstract explanatory facts. 
Conversely, the US model exhibits increasing performance, with a trend that is inverse. Short explanations, indeed, tend to include question-specific facts with low explanatory power. On the other hand, the longer the explanation, the higher the number of core scientific facts. Therefore, the decrease in MAP observed for the RS model is compensated by the Unification Score, since core scientific facts tend to form unification patterns across similar questions. This results demonstrate that the Unification Score has a crucial role in alleviating the semantic drift for the joint model (RS + US), resulting in a larger improvement on many-hops explanations (6+ facts).

Similarly, Figure~\ref{fig:precision_k} illustrates the Precision@K. As shown in the graph, the drop in precision for the US model exhibits the slowest degradation. Similarly to what observed for many-hops explanations, the US score contributes to the robustness of the RS + US model, making it able to reconstruct more precise explanations. As discussed in section \ref{sec:answ_sel}, this feature has a positive impact on question answering.

\paragraph{k-NN clustering.} We investigate the impact of the k-NN clustering on the explanation reconstruction task. Figure \ref{fig:impact_knn} shows the MAP score obtained by the joint RS + US model (BM25) with different numbers $k$ of nearest hypotheses considered for the Unification Score. The graph highlights the improvement in MAP achieved with increasing values of $k$. Specifically, we observe that the best MAP is obtained with $k=100$. These results confirm that the explanatory power can be effectively estimated using clusters of similar hypotheses, and that the unification-based mechanism has a crucial role in improving the performance of the relevance model.

\subsection{Qualitative analysis.}
To provide additional insights on the complementary aspects of Unification and Relevance Score, we present a set of qualitative examples from the dev-set. Table \ref{tab:qualitative_examples} illustrates the ranking assigned by RS and RS + US models to scientific sentences of increasing complexity. The words in \textbf{bold} indicate lexical overlaps between question, answer and explanation sentence. In the first example, the sentence \emph{``gravity; gravitational force causes objects  that  have  mass;  substances  to  be pulled down; to fall on a planet''} shares key terms with question and candidate answer and is therefore relatively easy to rank for the RS model (\#36). Nevertheless, the RS + US model is able to improve the ranking by 34 positions (\#2), as the gravitational law represents a scientific pattern with high explanatory unification, frequently reused across similar questions. The impact of the Unification Score is more evident when considering abstract explanatory facts. Coming back to our original example (i.e. \emph{``What is an example of a force producing heat?''}), the fact \emph{``friction causes the temperature of an object to increase''} has no significant overlaps with question and answer. Thus, the RS model ranks the gold explanation sentence in a low position (\#1472). However, the Unification Score (US) is able to capture the explanatory power of the fact from similar hypotheses in $E_{kb}$, pushing the RS + US ranking up to position \#21 (+1451).

\subsection{Question Answering}

To understand whether the constructed  explanations can support question answering, we compare the performance of BERT for multiple-choice QA ~\cite{devlin2019bert} without explanations with the performance of BERT provided with the top K explanation sentences retrieved by RS and RS + US models (BM25). BERT without explanations operates on question and candidate answer only. On the other hand,
BERT with explanation receives the following input: the question ($q$), a candidate answer ($c_i$) and the explanation for $c_i$ ($E_i$).
In this setting, the model is fine-tuned for binary classification ($bert_b$) to predict a set of probability scores  $P = \{p_1,~p_2,~...,~p_n\}$ for each candidate answer in $C = \{c1,~c_2,~...,~c_n\}$:
\begin{equation}
    bert_b(\textrm{\texttt{[CLS]}}~||~q||c_i~||~\textrm{\texttt{[SEP]}}~||~E_{i}) = p_i
~\label{eq:bert_answer}
\end{equation}
The binary classifier operates on the final hidden state corresponding to the \texttt{[CLS]} token.
To answer the question $q$, the model selects the candidate answer $c_a$ such that $a = \argmax_i p_i$.

\label{sec:answ_sel}
\begin{table}[t]
    \ra{1}
    \small
    \centering
    \begin{tabular}{@{}p{3.5cm}ccc@{}}
    \toprule
         \multirow{2}{*}{\textbf{Model}} &
         \multicolumn{3}{c}{\textbf{Accuracy}}\\
         \cmidrule{2-4}
         &Easy&Challenge&Overall\\
         \midrule
         BERT (no explanation)& 48.54 & 26.28 & 41.78\\
         \midrule
         BERT + RS (K = 3) & 53.20 & 40.97 &  49.39\\
         BERT + RS (K = 5) & 54.36 & 38.14 & 49.31\\
         BERT + RS (K = 10) & 32.71 & 29.63 & 31.75\\
         \midrule
         BERT + RS + US (K = 3) & \textbf{55.46} & \textbf{41.97} &  \textbf{51.62} \\
         BERT + RS + US (K = 5) & 54.48  & 39.43 & 50.12\\
         BERT + RS + US (K = 10) & 48.66 & 37.37 & 45.14\\
         \bottomrule
    \end{tabular}
    \caption{Performance of BERT on question answering (test-set) with and without the explanation reconstruction models.}
    \label{tab:bert_ans_selection}
\end{table}

Table~\ref{tab:bert_ans_selection}  reports the accuracy with and without explanations on the Worldtree \emph{test-set} for \emph{easy} and \emph{challenge} questions~\cite{clark2018think}.
Notably, a significant improvement in accuracy can be observed when BERT is provided with explanations retrieved by the reconstruction modules (+9.84\% accuracy with RS BM25 + US BM25 model). The improvement is consistent on the \emph{easy} split (+6.92\%) and particularly significant for \emph{challenge} questions (+15.69\%). Overall, we observe a correlation between more precise explanations and accuracy in answer prediction, with BERT + RS being outperformed by BERT + RS + US for each value of K. The decrease in accuracy occurring with increasing values of K is coherent with the drop in precision for the models observed in Figure~\ref{fig:precision_k}.
Moreover, steadier results adopting the RS + US model suggest a positive contribution from abstract explanatory facts. 
Additional investigation of this aspect will be a focus for future work.

\section{Conclusion}
This paper proposed a novel framework for multi-hop explanation reconstruction based on \emph{explanatory unification}. An extensive evaluation on the Worldtree corpus led to the following conclusions: (1) The approach is competitive with state-of-the-art Transformers, yet being significantly faster and inherently scalable; (2) The unification-based mechanism supports the construction of complex and many hops explanations; (3) The constructed explanations improves the accuracy of BERT for question answering  by up to 10\% overall. As a future work, we plan to extend the framework adopting neural embeddings for sentence representation.

\section*{Acknowledgements}

The authors would like to thank the anonymous reviewers for the constructive feedback.  A special thanks to Deborah Ferreira for the helpful discussions, and to the members of the AI Systems lab from the University of Manchester. Additionally, we would like to thank the Computational Shared Facility of the University of Manchester for providing the infrastructure to run our experiments.

\bibliography{eacl2021}

\begin{thebibliography}{43}
\expandafter\ifx\csname natexlab\endcsname\relax\def\natexlab#1{#1}\fi

\bibitem[{Banerjee(2019)}]{banerjee2019asu}
Pratyay Banerjee. 2019.
\newblock Asu at textgraphs 2019 shared task: Explanation regeneration using
  language models and iterative re-ranking.
\newblock In \emph{Proceedings of the Thirteenth Workshop on Graph-Based
  Methods for Natural Language Processing (TextGraphs-13)}, pages 78--84.

\bibitem[{Biran and Cotton(2017)}]{biran2017explanation}
Or~Biran and Courtenay Cotton. 2017.
\newblock Explanation and justification in machine learning: A survey.
\newblock In \emph{IJCAI-17 workshop on explainable AI (XAI)}, volume~8.

\bibitem[{Camburu et~al.(2018)Camburu, Rockt{\"a}schel, Lukasiewicz, and
  Blunsom}]{camburu2018snli}
Oana-Maria Camburu, Tim Rockt{\"a}schel, Thomas Lukasiewicz, and Phil Blunsom.
  2018.
\newblock e-snli: Natural language inference with natural language
  explanations.
\newblock In \emph{Advances in Neural Information Processing Systems}, pages
  9539--9549.

\bibitem[{Chia et~al.(2019)Chia, Witteveen, and Andrews}]{chia2019red}
Yew~Ken Chia, Sam Witteveen, and Martin Andrews. 2019.
\newblock Red dragon ai at textgraphs 2019 shared task: Language model assisted
  explanation generation.
\newblock In \emph{Proceedings of the Thirteenth Workshop on Graph-Based
  Methods for Natural Language Processing (TextGraphs-13)}, pages 85--89.

\bibitem[{Clark et~al.(2018)Clark, Cowhey, Etzioni, Khot, Sabharwal, Schoenick,
  and Tafjord}]{clark2018think}
Peter Clark, Isaac Cowhey, Oren Etzioni, Tushar Khot, Ashish Sabharwal, Carissa
  Schoenick, and Oyvind Tafjord. 2018.
\newblock Think you have solved question answering? try arc, the ai2 reasoning
  challenge.
\newblock \emph{arXiv preprint arXiv:1803.05457}.

\bibitem[{Das et~al.(2019)Das, Godbole, Zaheer, Dhuliawala, and
  McCallum}]{das2019chains}
Rajarshi Das, Ameya Godbole, Manzil Zaheer, Shehzaad Dhuliawala, and Andrew
  McCallum. 2019.
\newblock Chains-of-reasoning at textgraphs 2019 shared task: Reasoning over
  chains of facts for explainable multi-hop inference.
\newblock In \emph{Proceedings of the Thirteenth Workshop on Graph-Based
  Methods for Natural Language Processing (TextGraphs-13)}, pages 101--117.

\bibitem[{De~Mantaras et~al.(2005)De~Mantaras, McSherry, Bridge, Leake, Smyth,
  Craw, Faltings, Maher, T~COX, Forbus et~al.}]{de2005retrieval}
Ramon~Lopez De~Mantaras, David McSherry, Derek Bridge, David Leake, Barry
  Smyth, Susan Craw, Boi Faltings, Mary~Lou Maher, MICHAEL T~COX, Kenneth
  Forbus, et~al. 2005.
\newblock Retrieval, reuse, revision and retention in case-based reasoning.
\newblock \emph{The Knowledge Engineering Review}, 20(3):215--240.

\bibitem[{Devlin et~al.(2019)Devlin, Chang, Lee, and
  Toutanova}]{devlin2019bert}
Jacob Devlin, Ming-Wei Chang, Kenton Lee, and Kristina Toutanova. 2019.
\newblock Bert: Pre-training of deep bidirectional transformers for language
  understanding.
\newblock In \emph{Proceedings of the 2019 Conference of the North American
  Chapter of the Association for Computational Linguistics: Human Language
  Technologies, Volume 1 (Long and Short Papers)}, pages 4171--4186.

\bibitem[{D’Souza et~al.(2019)D’Souza, Mulang, and Auer}]{d2019team}
Jennifer D’Souza, Isaiah~Onando Mulang, and S{\"o}ren Auer. 2019.
\newblock Team svmrank: Leveraging feature-rich support vector machines for
  ranking explanations to elementary science questions.
\newblock In \emph{Proceedings of the Thirteenth Workshop on Graph-Based
  Methods for Natural Language Processing (TextGraphs-13)}, pages 90--100.

\bibitem[{Ferreira and Freitas(2020{\natexlab{a}})}]{ferreira2020natural}
Deborah Ferreira and Andr{\'e} Freitas. 2020{\natexlab{a}}.
\newblock Natural language premise selection: Finding supporting statements for
  mathematical text.
\newblock In \emph{Proceedings of The 12th Language Resources and Evaluation
  Conference}, pages 2175--2182.

\bibitem[{Ferreira and Freitas(2020{\natexlab{b}})}]{ferreira2020premise}
Deborah Ferreira and Andr{\'e} Freitas. 2020{\natexlab{b}}.
\newblock Premise selection in natural language mathematical texts.
\newblock In \emph{Proceedings of the 58th Annual Meeting of the Association
  for Computational Linguistics}, pages 7365--7374.

\bibitem[{Fried et~al.(2015)Fried, Jansen, Hahn-Powell, Surdeanu, and
  Clark}]{fried2015higher}
Daniel Fried, Peter Jansen, Gustave Hahn-Powell, Mihai Surdeanu, and Peter
  Clark. 2015.
\newblock Higher-order lexical semantic models for non-factoid answer
  reranking.
\newblock \emph{Transactions of the Association for Computational Linguistics},
  3:197--210.

\bibitem[{Friedman(1974)}]{friedman1974explanation}
Michael Friedman. 1974.
\newblock Explanation and scientific understanding.
\newblock \emph{The Journal of Philosophy}, 71(1):5--19.

\bibitem[{Hempel(1965)}]{hempel1965aspects}
Carl~G Hempel. 1965.
\newblock Aspects of scientific explanation.

\bibitem[{Jansen et~al.(2016)Jansen, Balasubramanian, Surdeanu, and
  Clark}]{jansen2016s}
Peter Jansen, Niranjan Balasubramanian, Mihai Surdeanu, and Peter Clark. 2016.
\newblock What’s in an explanation? characterizing knowledge and inference
  requirements for elementary science exams.
\newblock In \emph{Proceedings of COLING 2016, the 26th International
  Conference on Computational Linguistics: Technical Papers}, pages 2956--2965.

\bibitem[{Jansen et~al.(2017)Jansen, Sharp, Surdeanu, and
  Clark}]{jansen2017framing}
Peter Jansen, Rebecca Sharp, Mihai Surdeanu, and Peter Clark. 2017.
\newblock Framing qa as building and ranking intersentence answer
  justifications.
\newblock \emph{Computational Linguistics}, 43(2):407--449.

\bibitem[{Jansen and Ustalov(2019)}]{jansen2019textgraphs}
Peter Jansen and Dmitry Ustalov. 2019.
\newblock Textgraphs 2019 shared task on multi-hop inference for explanation
  regeneration.
\newblock In \emph{Proceedings of the Thirteenth Workshop on Graph-Based
  Methods for Natural Language Processing (TextGraphs-13)}, pages 63--77.

\bibitem[{Jansen et~al.(2018)Jansen, Wainwright, Marmorstein, and
  Morrison}]{jansen2018worldtree}
Peter Jansen, Elizabeth Wainwright, Steven Marmorstein, and Clayton Morrison.
  2018.
\newblock Worldtree: A corpus of explanation graphs for elementary science
  questions supporting multi-hop inference.
\newblock In \emph{Proceedings of the Eleventh International Conference on
  Language Resources and Evaluation (LREC 2018)}.

\bibitem[{Khashabi et~al.(2019)Khashabi, Azer, Khot, Sabharwal, and
  Roth}]{khashabi2019capabilities}
Daniel Khashabi, Erfan~Sadeqi Azer, Tushar Khot, Ashish Sabharwal, and Dan
  Roth. 2019.
\newblock On the capabilities and limitations of reasoning for natural language
  understanding.
\newblock \emph{arXiv preprint arXiv:1901.02522}.

\bibitem[{Khashabi et~al.(2018)Khashabi, Khot, Sabharwal, and
  Roth}]{khashabi2018question}
Daniel Khashabi, Tushar Khot, Ashish Sabharwal, and Dan Roth. 2018.
\newblock Question answering as global reasoning over semantic abstractions.
\newblock In \emph{Thirty-Second AAAI Conference on Artificial Intelligence}.

\bibitem[{Khot et~al.(2019)Khot, Clark, Guerquin, Jansen, and
  Sabharwal}]{khot2019qasc}
Tushar Khot, Peter Clark, Michal Guerquin, Peter Jansen, and Ashish Sabharwal.
  2019.
\newblock Qasc: A dataset for question answering via sentence composition.
\newblock \emph{arXiv preprint arXiv:1910.11473}.

\bibitem[{Khot et~al.(2017)Khot, Sabharwal, and Clark}]{khot2017answering}
Tushar Khot, Ashish Sabharwal, and Peter Clark. 2017.
\newblock Answering complex questions using open information extraction.
\newblock In \emph{Proceedings of the 55th Annual Meeting of the Association
  for Computational Linguistics (Volume 2: Short Papers)}, pages 311--316.

\bibitem[{Kitcher(1981)}]{kitcher1981explanatory}
Philip Kitcher. 1981.
\newblock Explanatory unification.
\newblock \emph{Philosophy of science}, 48(4):507--531.

\bibitem[{Kitcher(1989)}]{kitcher1989explanatory}
Philip Kitcher. 1989.
\newblock Explanatory unification and the causal structure of the world.

\bibitem[{Kolodner(2014)}]{kolodner2014case}
Janet Kolodner. 2014.
\newblock \emph{Case-based reasoning}.
\newblock Morgan Kaufmann.

\bibitem[{Lacave and Diez(2004)}]{lacave2004review}
Carmen Lacave and Francisco~J Diez. 2004.
\newblock A review of explanation methods for heuristic expert systems.
\newblock \emph{The Knowledge Engineering Review}, 19(2):133--146.

\bibitem[{Mihaylov et~al.(2018)Mihaylov, Clark, Khot, and
  Sabharwal}]{mihaylov2018can}
Todor Mihaylov, Peter Clark, Tushar Khot, and Ashish Sabharwal. 2018.
\newblock Can a suit of armor conduct electricity? a new dataset for open book
  question answering.
\newblock In \emph{Proceedings of the 2018 Conference on Empirical Methods in
  Natural Language Processing}, pages 2381--2391.

\bibitem[{Miller(2019)}]{miller2019explanation}
Tim Miller. 2019.
\newblock Explanation in artificial intelligence: Insights from the social
  sciences.
\newblock \emph{Artificial Intelligence}, 267:1--38.

\bibitem[{Mitchell et~al.(1986)Mitchell, Keller, and
  Kedar-Cabelli}]{mitchell1986explanation}
Tom~M Mitchell, Richard~M Keller, and Smadar~T Kedar-Cabelli. 1986.
\newblock Explanation-based generalization: A unifying view.
\newblock \emph{Machine learning}, 1(1):47--80.

\bibitem[{Pearl(2009)}]{pearl2009causality}
Judea Pearl. 2009.
\newblock \emph{Causality}.
\newblock Cambridge university press.

\bibitem[{Rajani et~al.(2019)Rajani, McCann, Xiong, and
  Socher}]{rajani2019explain}
Nazneen~Fatema Rajani, Bryan McCann, Caiming Xiong, and Richard Socher. 2019.
\newblock Explain yourself! leveraging language models for commonsense
  reasoning.
\newblock In \emph{Proceedings of the 57th Annual Meeting of the Association
  for Computational Linguistics}, pages 4932--4942.

\bibitem[{Ribeiro et~al.(2016)Ribeiro, Singh, and Guestrin}]{ribeiro2016should}
Marco~Tulio Ribeiro, Sameer Singh, and Carlos Guestrin. 2016.
\newblock " why should i trust you?" explaining the predictions of any
  classifier.
\newblock In \emph{Proceedings of the 22nd ACM SIGKDD international conference
  on knowledge discovery and data mining}, pages 1135--1144.

\bibitem[{Robertson et~al.(2009)Robertson, Zaragoza
  et~al.}]{robertson2009probabilistic}
Stephen Robertson, Hugo Zaragoza, et~al. 2009.
\newblock The probabilistic relevance framework: Bm25 and beyond.
\newblock \emph{Foundations and Trends{\textregistered} in Information
  Retrieval}, 3(4):333--389.

\bibitem[{Salmon(1984)}]{salmon1984scientific}
Wesley~C Salmon. 1984.
\newblock \emph{Scientific explanation and the causal structure of the world}.
\newblock Princeton University Press.

\bibitem[{S{\o}rmo et~al.(2005)S{\o}rmo, Cassens, and
  Aamodt}]{sormo2005explanation}
Frode S{\o}rmo, J{\"o}rg Cassens, and Agnar Aamodt. 2005.
\newblock Explanation in case-based reasoning--perspectives and goals.
\newblock \emph{Artificial Intelligence Review}, 24(2):109--143.

\bibitem[{Thagard(1992)}]{thagard1992analogy}
Paul Thagard. 1992.
\newblock Analogy, explanation, and education.
\newblock \emph{Journal of research in science teaching}, 29(6):537--544.

\bibitem[{Thagard and Litt(2008)}]{thagard2008models}
Paul Thagard and Abninder Litt. 2008.
\newblock Models of scientific explanation.
\newblock \emph{The Cambridge handbook of computational psychology}, pages
  549--564.

\bibitem[{Thayaparan et~al.(2020)Thayaparan, Valentino, and
  Freitas}]{thayaparan2020survey}
Mokanarangan Thayaparan, Marco Valentino, and Andr{\'e} Freitas. 2020.
\newblock A survey on explainability in machine reading comprehension.
\newblock \emph{arXiv preprint arXiv:2010.00389}.

\bibitem[{Thayaparan et~al.(2019)Thayaparan, Valentino, Schlegel, and
  Freitas}]{thayaparan2019identifying}
Mokanarangan Thayaparan, Marco Valentino, Viktor Schlegel, and Andr{\'e}
  Freitas. 2019.
\newblock Identifying supporting facts for multi-hop question answering with
  document graph networks.
\newblock In \emph{Proceedings of the Thirteenth Workshop on Graph-Based
  Methods for Natural Language Processing (TextGraphs-13)}, pages 42--51.

\bibitem[{Wick and Thompson(1992)}]{wick1992reconstructive}
Michael~R Wick and William~B Thompson. 1992.
\newblock Reconstructive expert system explanation.
\newblock \emph{Artificial Intelligence}, 54(1-2):33--70.

\bibitem[{Woodward(2005)}]{woodward2005making}
James Woodward. 2005.
\newblock \emph{Making things happen: A theory of causal explanation}.
\newblock Oxford university press.

\bibitem[{Yadav et~al.(2019)Yadav, Bethard, and Surdeanu}]{yadav2019quick}
Vikas Yadav, Steven Bethard, and Mihai Surdeanu. 2019.
\newblock Quick and (not so) dirty: Unsupervised selection of justification
  sentences for multi-hop question answering.
\newblock In \emph{Proceedings of the 2019 Conference on Empirical Methods in
  Natural Language Processing and the 9th International Joint Conference on
  Natural Language Processing (EMNLP-IJCNLP)}, pages 2578--2589.

\bibitem[{Yang et~al.(2018)Yang, Qi, Zhang, Bengio, Cohen, Salakhutdinov, and
  Manning}]{yang2018hotpotqa}
Zhilin Yang, Peng Qi, Saizheng Zhang, Yoshua Bengio, William Cohen, Ruslan
  Salakhutdinov, and Christopher~D Manning. 2018.
\newblock Hotpotqa: A dataset for diverse, explainable multi-hop question
  answering.
\newblock In \emph{Proceedings of the 2018 Conference on Empirical Methods in
  Natural Language Processing}, pages 2369--2380.

\end{thebibliography}
\bibliographystyle{acl_natbib}

\appendix

\section{Supplementary Material}

\subsection{Hyperparameters tuning}
The hyperparameters of the model have been tuned manually. The criteria for the optimisation is the maximisation of the MAP score on the dev-set. Here, we report the values adopted for the experiments described in the paper.

The Unification-based Reconstruction adopts two hyperparameters. Specifically, $\lambda_1$ is  the weight assigned to the relevance score in equation \ref{eq:model_combination}, while $k$ is the number of similar hypotheses to consider for the calculation of the unification score (equation \ref{eq:unification_score}). The values adopted for these parameters are as follows:
\begin{enumerate}
    \item $\lambda_1$ = 0.83 ($ 1 - \lambda_1$ = 0.17)
    \item $k$ = 100
\end{enumerate}

\subsection{BERT model}

For question answering we adopt a BERT$_{BASE}$ model. The model is implemented using PyTorch (\url{https://pytorch.org/}) and fine-tuned using 4 Tesla 16GB V100 GPUs for 10 epochs in total with batch size 32 and seed 42.
The hyperparameters adopted for BERT are as follows:
\begin{itemize}
    \item gradient\_accumulation\_steps = 1
    \item learning\_rate = 5e-5
    \item weight\_decay = 0.0
    \item adam\_epsilon = 1e-8
    \item warmup\_steps = 0
    \item max\_grad\_norm = 1.0
\end{itemize}

\subsection{Data and code}
The experiments are carried out on the TextGraphs 2019 version (\url{https://github.com/umanlp/tg2019task}) of the Worldtree corpus. The full dataset can be downloaded at the following URL: \url{http://cognitiveai.org/dist/worldtree_corpus_textgraphs2019sharedtask_withgraphvis.zip}.

The code to reproduce the experiments described in the paper is available at the following URL: \url{https://github.com/ai-systems/unification_reconstruction_explanations}

\end{document}